\documentclass{article}

\usepackage[nonatbib]{neurips_2022}

\usepackage{multirow}
\usepackage[utf8]{inputenc} 
\usepackage[T1]{fontenc}    
\usepackage{hyperref}       
\usepackage{url}            
\usepackage{booktabs}       
\usepackage{amsfonts}       
\usepackage{nicefrac}       
\usepackage{microtype}      
\usepackage{xcolor}         
\usepackage{natbib}
\usepackage{amsmath}
\usepackage{amssymb}
\usepackage{mathtools}
\usepackage{amsthm}
\usepackage{graphicx}  
\usepackage{float}
\usepackage{bbm}
\usepackage{siunitx}
\usepackage{booktabs}
\usepackage{caption}
\usepackage{changepage}
\usepackage{tikz}
\usepackage[symbol]{footmisc}
\usepackage[T1]{fontenc}
\hypersetup{hidelinks}

\usetikzlibrary{positioning}
\usetikzlibrary{decorations.pathreplacing}

\title{Learning high-dimensional causal effect}

\author{%
  Aayush Agarwal$^*$ \\
  Courant Institute of Mathematical Sciences\\
  New York University\\
  \texttt{aka7919@nyu.edu} \\
  \And
  Saksham Bassi\thanks{Equal contribution} \\
  Courant Institute of Mathematical Sciences\\
  New York University\\
  \texttt{sakshambassi@nyu.edu} \\
}

\begin{document}

\maketitle

\begin{abstract}
The scarcity of high-dimensional causal inference datasets restricts the exploration of complex deep models. In this work, we propose a method to generate a synthetic causal dataset that is high-dimensional. The synthetic data simulates a causal effect using the MNIST dataset with Bernoulli treatment values. This provides an opportunity to study varieties of models for causal effect estimation. We experiment on this dataset using Dragonnet architecture (\cite{dragonnet}) and modified architectures. We use the modified architectures to explore different types of initial Neural Network layers and observe that the modified architectures perform better in estimations. We observe that residual and transformer models estimate treatment effect very closely without the need for targeted regularization, introduced by \cite{dragonnet}.\footnote[2]{Please find the code for this paper here: \url{https://github.com/sakshambassi/HD-Causal-Effect}}
\end{abstract}

\section{Introduction}

\label{intro}
High-dimensional datasets enable testing state-of-the-art models for various new tasks. When a dataset is high dimensional, it opens the opportunity of experimenting with different kinds of deep learning models. Therefore, the solution is to acquire and develop high-dimensional data like images or human language and synthesize a dataset based on the same for causal inference.

Deep Learning models can learn patterns without prior underlying information. Work done by \cite{tarnet} was crucial as it applies the ability of learning of treatment estimation, specifically for Individual Treatment Effect estimation. Individual Treatment Effect  is basically quantifying the effect of an intervention (treatment) on the outcome of individual input. It is formulated as $\tau(X) = E[Y|X,T=1] - E[Y | X, T=0]$, where $X$ is the input, $Y$ is the outcome and $T$ is the treatment.

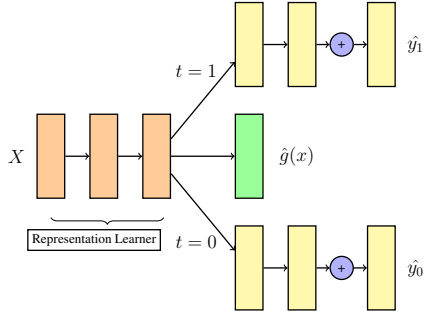
\begin{figure}[t]
    \centering
    \scalebox{0.5}{
         \begin{tikzpicture}[node distance=7em, every node/.style = {shape=rectangle, draw,
                             minimum height=1.5em, line width=1pt,
                             align=center}, every edge/.append style = {line width=1pt}]
             \node[text width=5mm, text height=20mm, fill=orange!40] (f1) {};
             
             \node[right of=f1, text width=5mm,text height=20mm, xshift=-3em, fill=orange!40] (f2) {};
             
             \node[right of=f2, text width=5mm,text height=20mm, xshift=-3em, fill=orange!40] (f3){};
             
             \node[right of=f3, above=2em of f3,text height=20mm, text width=5mm, fill=yellow!40] (c1){};
             
             \node[right of=c1, text width=5mm,text height=20mm, xshift=-3em, fill=yellow!40] (c2){};
             
             \node[right of=c1, shape=circle, text width=2mm, text height=2mm, fill=blue!30] (con) {+};
             
             \node[right of=con, text width=5mm,text height=20mm, xshift=-4em, fill=yellow!40] (c3){};
             
             \node[right of=f3, text width=5mm,text height=20mm, fill=green!40] (g1){};
             
             \node[right of=f3, below=2em of f3,text height=20mm, text width=5mm, fill=yellow!40] (d1){};
             
             \node[right of=d1, text width=5mm,text height=20mm, xshift=-3em, fill=yellow!40] (d2){};
             
             \node[right of=d1, shape=circle, text width=2mm, text height=2mm, fill=blue!30] (con2) {+};
             
             \node[right of=con2, text width=5mm,text height=20mm, xshift=-4em, fill=yellow!40] (d3){};
             
             \draw [decorate,decoration={brace,amplitude=5pt, mirror, raise=4ex}]
             (0,-1) -- (3,-1) node[midway,yshift=-3.5em,xshift=-1em]{Representation Learner};
             \draw [->] (f1) edge (f2);
             \draw [->] (f2) edge (f3);
             \draw [->] (f3) edge (g1);
             \draw [->] (f3) edge (c1);
             \draw [->] (f3) edge (d1);
             \draw [->] (c1) edge (c2);
             \draw [->] (c2) edge (con);
             \draw [->] (con) edge (c3);
             \draw [->] (d1) edge (d2);
             \draw [->] (d2) edge (con2);
             \draw [->] (con2) edge (d3);
             
             \node[left of=f1, node distance=2.6em, yshift=0em, draw=none] {\Large $X$};
             
             \node[right of=g1, node distance=3.6em, yshift=0em, draw=none] {\Large $\hat{g}(x)$};
             
             \node[right of=c3, node distance=2.6em, yshift=0em, draw=none] {\Large $\hat{y_1}$};
             
             \node[right of=d3, node distance=2.6em, yshift=0em, draw=none] {\Large $\hat{y_0}$};
             
             \node[left of=c1, node distance=4.0em, yshift=-2.em, draw=none] {\Large ${t=1}$};
             
             \node[left of=d1, node distance=4.0em, yshift=+2.em, draw=none] {\Large ${t=0}$};
             
         \end{tikzpicture}
    }
    \caption{Dragonnet architecture proposed in \cite{dragonnet}. $\hat{y_1}$ and $\hat{y_0}$ are outcome heads. $\hat{g}(x)$ is the estimation head of propensity score.}
    \label{fig:concat_dragonnet}
\end{figure}

Along with individual treatment effect, we have Average Treatment Effect(ATE), which is formulated as $E\Big[E[Y|X,T=1] - E[Y|X,T=0]\Big]$ or $E[\tau(x)]$, where $\tau$ is the individual treatment effect. When simulating a dataset, we can compute the ATE of true data distribution. This opens the possibility of comparing ATE of the estimator model with that of the true data distribution to quantify how good an estimator model estimates the treatment effect. This would serve as one of the metrics for our experiments. A similar comparison can be done on the direct outcome variable estimation which is common in regular regression or classification problems. When it is required to analyze treatment effect on the treated only, we use Average Treatment Effect on Treated (ATT) i.e. formulated as $E\Big[E[Y_1 - Y_0 | T = 1]\Big]$. Here, $Y_1$ is the outcome if treatment were applied to the input, and $Y_0$ is the outcome if treatment were not applied to the input. Another important function of $x$ is propensity score represented as $g(X)$ is $P(T=1|X)$. The propensity score is the probability of input to receive treatment. \citeauthor{tarnet} introduced a deep learning model to estimate treatment effects that had an architecture based on deep network layers with two outcome heads which output $Y_1$ and $Y_0$. \citeauthor{dragonnet} proposed an architecture called Dragonnet which was inspired by \cite{tarnet} with a modification of adding propensity score estimation while estimating target values $Y$. The Dragonnet model saw improvements in the estimation quality.

In this work, we study the following:
\begin{itemize}
    \item First, we propose a method to simulate a high dimensional causal effect dataset based on MNIST images to enable experiments with a variety of deep learning models.

    \item Second, we experiment with different initial layers (we call them representation learners, see Figure \ref{fig:concat_dragonnet}) in Dragonnet. We notice that the learners we use perform better than the one used in the Dragonnet architecture for our simulated high dimensional dataset without targetted regularization objective (introduced in \citet{dragonnet}).
\end{itemize}

We discuss the setup more in detail in Section \ref{subsec:mnist} and results in Section \ref{sec:experiment}.

\section{Related Work}

Previously, in the domain of understanding causality, linear frameworks have been applied to non-linear data distribution to find causal direction such as in the work of \citet{Hoyer08}. \citeauthor{tarnet} brought a breakthrough in Treatment estimation by introducing a neural network architecture called TARNET. TARNET has similar architecture to Figure \ref{fig:concat_dragonnet} but without the propensity head estimation.  \citeauthor{Johansson16} introduced two methods to use representation learning on a set of covariates for counterfactual inference, out of which one method was a deep learning framework called Balancing Neural Network. Another work focusing on using Variation Auto-Encoders for causal effect estimation (independent and population causal effects) is proposed in \cite{Louizos17}. \citeauthor{cate} proposed that it is important to understand the data generating process and, thus, it is possible that some datasets favor some algorithms more than others. The paper discusses the performance of random forests and neural networks based on causal estimators on IHDP and ACIC datasets for inferring the same.

\citeauthor{dragonnet} proposed two things in their work, first a modified neural network architecture (called Dragonnet; based on TARNET \cite{tarnet}) and second, a targetted regularization technique as an objective function (called t-reg). The modification in Dragonnet is that they have added another estimation head for estimating propensity score $g(x)$. They studied that Dragonnet performed better in treatment estimation as the model itself selects an optimal tradeoff between the error in propensity score and outcome. \citeauthor{dragonnet} also discusses a dense network Nednet, which is trained with covariates $X$ as input and treatment $t$ as output. Then removing Nednet's final layers, the weights of the dense network are reused to output the potential outcomes $Y_0$ and $Y_1$. Later, the paper compares that the Dragonnet model is better than the Nednet model at handling the tradeoff between propensity score representation ($G(x)$) and accuracy of prediction ($Y$).

\begin{figure*}[t]
    \centering
    \scalebox{0.6}{
         \begin{tikzpicture}[node distance=7em, every node/.style = {shape=rectangle, draw,
                             minimum height=1.5em, line width=1pt,
                             align=center, text height=3mm}, every edge/.append style = {line width=1pt}]
             \node[text width=25mm, fill=cyan!30] (f1) {{$X$ (as Images)}};
             
             \node[right of=f1, text width=35mm, xshift=20em, fill=cyan!30] (f2) {{$X_{labels}^i \in X_{labels}$}};
             
             \node[below=2.5em of f2, text width=25mm, xshift=6em, fill=green!30] (f3) {$Bern(p=0.8)$};
             
             \node[below=2.5em of f2, text width=25mm, xshift=-6
             em, fill=green!30] (f4) {$Bern(p=0.2)$};
             
             \node[below=2em of f4, text width=35mm, fill=blue!30] (f5) {\textbf{$t$: Binary variable}};
             
             \node[below=2em of f3, text width=35mm, fill=blue!30] (f6) {\textbf{$t$: Binary variable}};
             
             \node[below=1.5em of f1, text width=55mm, fill=green!30] (c1) {Noisy CNN + FC layer + Log Softmax layer};
             
             \node[below=1.5em of c1, text width=35mm, fill=orange!40] (d1) {{Array of $[p_1,p_2,p_3,...,p_{10}]$}};

             \node[below=3em of d1, text width=45mm, text height=1em, fill=green!30] (e1) {$Bern(p=arg\_max)$};
             
             \node[below=2.5em of e1, text width=65mm, fill=blue!25] (e2) {\textbf{$Y_0$ : Array of binary variable}};
             
             \node[below=0.5em of d1, node distance=1.6em, xshift=-2.5em, yshift=0em, draw=none] {{$arg\_max$}};
             
             \node[below=0.5em of e1, node distance=1.6em, xshift=-2.5em, yshift=0em, draw=none] {{Sampling}};
             
             \node[below=0.1em of f2, node distance=1.6em, xshift=-7.0em, yshift=0em, draw=none] {{if $X_{labels}^{i}$ is even}};
             
             \node[below=0.1em of f2, node distance=1.6em, xshift=7.0em, yshift=0em, draw=none] {{if $X_{labels}^{i}$ is odd}};
             
             \draw [->] (f1) edge (c1);
             \draw [->] (c1) edge (d1);
             \draw [->] (e1) edge (e2);
        
            \draw [->] (d1) edge (e1);
            \draw [->] (f2) edge (f3);
            \draw [->] (f2) edge (f4);
            \draw [->] (f4) edge (f5);
            \draw [->] (f3) edge (f6);
         \end{tikzpicture}
    }
    \caption{Simulation process for synthetic dataset creation based on MNIST and few Bernoulli distributions. Just like we have shown in the figure simulation of $Y_0$, we repeat this process with slightly different model parameters (in terms of Dropout values and Number of neurons) CNN for simulating $Y_1$.}
    \label{fig:simulation}
\end{figure*}
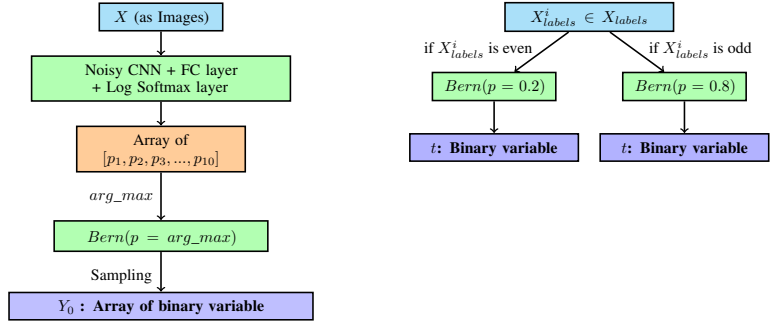

\section{Data \& Methodology}

In this section, we first describe the IHDP dataset which is used by \citeauthor{dragonnet} in their work. Following this, the simulation process using the MNIST images (\cite{lecun-mnist}) briefed above, is elaborated. The last part in this section explains the modified architectures we experimented in this work.

\subsection{IHDP dataset simulation by Dragonnet}

\cite{dragonnet} uses 50 replications of the IHDP dataset (with 747 observations each). Each replication consists of treatment $t \in \{0, 1\}$, $\mu_0$ and $\mu_1$ (where $\mu_0 = E[Y_0|X]$ and $\mu_1 = E[Y_1|X]$) as simulated outputs from different distributions and the above mentioned covariates, with 19 of them as binary values and 6 continuous values. From these $treatment$ and $\mu$ values, the paper computes a truth value for average treatment effect by calculating $\mathbb{E}\left[\mu_1 - \mu_0\right]$. This helps to measure the error over the estimated average treatment effect with respect to this true value.

\subsection{Simulated MNIST dataset}
\label{subsec:mnist}

This module includes simulating a causal effect for high dimensional attributes such as images as input $X$. We take the MNIST digits dataset (\cite{lecun-mnist}) for the experiment. Similar to the simulation process of the IHDP dataset above, we create attributes like $Y_0$, $Y_1$, $X$, and $t$. We create the different distributions for $Y_0$ and $Y_1$ by using a different combination of random seed and dropouts in a Convolutional Neural Network (CNN). The process of this creation is by using a noisy CNN with Log Softmax outputs, finding the maximum of the outputs, and passing it as a probability value to a Bernoulli distribution to sample a single output value. We use this process for creating a set of both $Y_0$ and $Y_1$.

The treatment $t$ is created by sampling from different Bernoulli distributions based on whether the label for the input image in the MNIST dataset is odd or even. If the label is odd, we sample $t$ from a Bernoulli distribution with a probability of getting a 1 equal to $0.8$. If the label is even, we sample $t$ from a Bernoulli distribution with a probability of getting a 1 equal to $0.2$. The true $Y$ is picked as either $Y_0$ or $Y_1$ depending on whether the treatment $t$ for the given sample is $0$ or $1$. We demonstrated the simulation process in Figure \ref{fig:simulation}.

Like in the simulated IHDP dataset above, we can now calculate the true Average Treatment Effect value, which will act as a reference to calculate the error (absolute difference) on the Average Treatment Effect estimated by the model.

\subsection{Representation Learners}

As discussed in the introduction (Section \ref{intro}), a high dimension input allows us to experiment with different types of encoders (representation learners, see Figure \ref{fig:concat_dragonnet}) to represent the image embedding. We use dense network from Dragonnet, ResNet50 (\cite{Kaiming16}) and ViT (\cite{vit}) as encoders in our experiment. When training the model, we hide the potential values $Y_0$ and $Y_1$ that we simulated and present the model with $X$, $Y$, and $t$ as the objective function only considers the outcome head estimate depending on $t$ and calculates loss based on the true values of $Y$. The results for the same can be found in Section \ref{sec:experiment}.

The targetted regularization presented in the work of \cite{dragonnet} is focused on a continuous variable as an outcome. However, since in our experiment, we simulate a dataset with the final outcome having a binary variable, we modify the regularization to include binary cross-entropy of the perturbation instead of squared error of the perturbation.

\subsubsection{ResNet as Representation Learner}

Increasing the dimensionality of input data facilitates making the first few layers of the network (encoder) pluggable. One of the additional encoders we experiment with is a pretrained ResNet50 (\cite{Kaiming16}). As the depth of the neural network increases, its number of parameters and hence its capacity to model complex distributions increases. However, it was observed that an $18$ layer deep model had a lower error than a $34$ layer deep network in the work of \cite{Kaiming16}. These counterintuitive observations could be attributed to the vanishing of gradients as loss is propagated backward into the layers. ResNet proposed skip connections between convolution layers. The advantage of this skip connection is that in a forward pass, the sample can choose whether it needs to be transformed by a complex convolution or simpler linearity. As a result, when the model learns the weights of the main trunk and skip connections, it effectively also chooses the number of convolutions in the model as a parameter, thus making the depth of the network flexible as per the dataset modeled.

To use it as transfer learning, we shave off the classification head of the ResNet model and passed the feature map of the image to the two outcome heads and to the propensity score estimation head as shown in Figure \ref{fig:concat_dragonnet}.

\subsubsection{Vision Transformer as Representation Learner}

Transformers (\cite{transformer}) have been used for NLP tasks previously which resulted in better accuracy and parallelization. Recently, the use of transformers (\cite{keras_vit}) has been significant in vision tasks as well. Similar to the NLP tasks, the images are partitioned into patches (which can be overlapping) that act as tokens to the transformer. Before passing it to attention heads, the embeddings for each of the patch is computed with a shared neural network, and the embeddings are then fused with a representative of the positional encoding of that particular patch with respect to the original image.

We shave off the classification head at the end like in the ResNet experiment and use it as a representation learner as shown in Figure \ref{fig:concat_dragonnet}. For Vision Transformer (ViT) network, we adapted code from \cite{keras_vit}.

\section{Experiment}

\label{sec:experiment}

Below are the formulae for the metrics calculated in Table \ref{tab:mnist_test}. We observe the metrics for absolute error (absolute difference) on average treatment effect (called ATE\textsubscript{AE}). As described in the previous section, the ATE of the true data distribution is used as a true reference for ATE\textsubscript{AE}.

{\small
\begin{align*}
    ATE\textsubscript{AE} &= \left|\mathbb{E}\left[Y_1 - Y_0\right] - \mathbb{E}\left[\hat{Y}_1 - \hat{Y}_0\right]\right| 
\end{align*}
}%

Here, $\hat{Y}_1, \hat{Y}_0$ are model estimates for $Y_1, Y_0$ respectively.

Our experiment is performed on the simulated dataset as shown in Figure \ref{fig:simulation}. We use Dragonnet as a baseline model and as discussed ResNet50 and ViT as representation learners for the architecture in Figure \ref{fig:concat_dragonnet}. The only change in the neural network in Figure \ref{fig:concat_dragonnet} is that we now have ResNet and ViT as representation learners additionally for separate experiments. The results are shown in Table \ref{tab:mnist_test}. We know that (from \cite{causality-handbook}),
{\small
\begin{align*}
    \underbrace{\mathbb{E}\left[\mathbb{E}\left[Y \mid T = 1\right] - \mathbb{E}\left[Y \mid T = 0\right]\right]}_{\text{Association}} &= \underbrace{\mathbb{E}\left[\mathbb{E}\left[Y_1 - Y_0 \mid T = 1\right]\right]}_{\text{ATT}} \\ + \underbrace{\mathbb{E}\left[\mathbb{E}\left[Y_0 \mid T = 1\right] - \mathbb{E}\left[Y_0 \mid T = 0\right]\right]}_{\text{Pre-treatment Bias}} \tag{1}
\end{align*}
}%

As seen in the Table \ref{tab:pre_treatment_test}, the pre-treatment bias obtained is close to $0$. This is helpful because the association is dependent on outcome $Y$ which we know, and not on potential outcomes $Y_0$ and $Y_1$ and we cannot know both of them. Also, Equation $(3)$ highlights that both the groups, the treated and the controlled are comparable, which is a desirable property for causal inference.
\begin{align*}
    \underbrace{\mathbb{E}\left[\mathbb{E}\left[Y \mid T = 1\right] - \mathbb{E}\left[Y \mid T = 0\right]\right]}_{\text{Association}} &= \underbrace{\mathbb{E}\left[\mathbb{E}\left[Y_1 - Y_0 \mid T = 1\right]\right]}_{\text{ATT}} \tag{2}\\
    \therefore \mathbb{E}\left[Y_0 \mid T = 1\right] \approx \mathbb{E}\left[Y_0 \mid T = 0\right] \tag{3}\\
\end{align*}

\begin{table}[t]
\vskip 0.15in
\begin{center}
\begin{small}
\begin{sc}
\begin{tabular}{lSSSSSS}
\toprule
\multirow{0}{*}{Model} &
      \multicolumn{2}{c}{Accuracy} &
      \multicolumn{2}{c}{ATE\textsubscript{AE}} \\
      & {w/o t-reg} & {t-reg} & {w/o t-reg} & {t-reg} \\
      \midrule
    Dragonnet & 54.67 & 65.97 & 0.017 & 0.021 \\
    \textbf{ResNet50 + Dragon heads} & \textbf{55.81} & 63.50  & \textbf{0.008} & \textbf{0.014} \\
    \textbf{ViT + Dragon heads} & \textbf{55.09}  & 52.42  & \textbf{0.001} & 0.532   \\
\bottomrule
\end{tabular}
\end{sc}
\end{small}
\end{center}
\caption{Result metrics on simulated Held-out {test} dataset. The models ResNet50 + Dragon heads and ViT + Dragon heads are representation learners with Dragon Heads.}
\label{tab:mnist_test}
\end{table}

\begin{table}[t]
\vskip 0.15in
\begin{center}
\begin{small}
\begin{sc}
\begin{tabular}{lSSSSSS}
\toprule
\multirow{0}{*}{Model} &
      \multicolumn{2}{c}{Pre-treatment bias} \\
      & {w/o t-reg} & {t-reg} \\
      \midrule
    Dragonnet & -0.0078 & -0.0007 \\
    \textbf{ResNet50 + Dragon heads} & -0.0163 &  -0.0141 \\
    \textbf{ViT + Dragon heads} &-0.0230  &-0.0113    \\

\bottomrule
\end{tabular}
\end{sc}
\end{small}
\end{center}
\caption{Pre-treatment bias results on simulated test data.}
\label{tab:pre_treatment_test}
\end{table}

\section*{Conclusion}



Our work unfolded the opportunity to explore recent Deep Learning models like Transformers and Residual Networks to estimate treatment effects. We also studied the changes in estimations by modifying the Dragonnet architecture. To create a platform for the exploration of deep networks, we introduced a simulation process for causal effect dataset creation. The outcome of our experiments on the simulated dataset shows that ResNet50 + Dragon heads model and ViT + Dragon heads models perform decently on the metrics of estimation error and accuracy.

\bibliography{bib}

\begin{thebibliography}{12}
\providecommand{\natexlab}[1]{#1}
\providecommand{\url}[1]{\texttt{#1}}
\expandafter\ifx\csname urlstyle\endcsname\relax
  \providecommand{\doi}[1]{doi: #1}\else
  \providecommand{\doi}{doi: \begingroup \urlstyle{rm}\Url}\fi

\bibitem[Alves(2021)]{causality-handbook}
Alves, M.~F.
\newblock Introduction to causality, 2021.
\newblock \\
  \href{https://matheusfacure.github.io/python-causality-handbook/01-Introduction-To-Causality.html}
  {\nolinkurl{https://matheusfacure.github.io/python-causality-handbook/} \\
  \nolinkurl{01-Introduction-To-Causality.html}}.

\bibitem[Curth et~al.(2021)Curth, Svensson, Weatherall, and van~der
  Schaar]{cate}
Curth, A., Svensson, D., Weatherall, J., and van~der Schaar, M.
\newblock Really doing great at estimating {CATE}? a critical look at {ML}
  benchmarking practices in treatment effect estimation.
\newblock In \emph{Thirty-fifth Conference on Neural Information Processing
  Systems Datasets and Benchmarks Track (Round 2)}, 2021.
\newblock URL \url{https://openreview.net/forum?id=FQLzQqGEAH}.

\bibitem[Dosovitskiy et~al.(2021)Dosovitskiy, Beyer, Kolesnikov, Weissenborn,
  Zhai, Unterthiner, Dehghani, Minderer, Heigold, Gelly, Uszkoreit, and
  Houlsby]{vit}
Dosovitskiy, A., Beyer, L., Kolesnikov, A., Weissenborn, D., Zhai, X.,
  Unterthiner, T., Dehghani, M., Minderer, M., Heigold, G., Gelly, S.,
  Uszkoreit, J., and Houlsby, N.
\newblock An image is worth 16x16 words: Transformers for image recognition at
  scale.
\newblock In \emph{International Conference on Learning Representations}, 2021.
\newblock URL \url{https://openreview.net/forum?id=YicbFdNTTy}.

\bibitem[He et~al.(2016)He, Zhang, Ren, and Sun]{Kaiming16}
He, K., Zhang, X., Ren, S., and Sun, J.
\newblock Deep residual learning for image recognition.
\newblock In \emph{2016 IEEE Conference on Computer Vision and Pattern
  Recognition (CVPR)}, pp.\  770--778, 06 2016.
\newblock \doi{10.1109/CVPR.2016.90}.

\bibitem[Hoyer et~al.(2008)Hoyer, Janzing, Mooij, Peters, and
  Sch\"{o}lkopf]{Hoyer08}
Hoyer, P., Janzing, D., Mooij, J.~M., Peters, J., and Sch\"{o}lkopf, B.
\newblock Nonlinear causal discovery with additive noise models.
\newblock In Koller, D., Schuurmans, D., Bengio, Y., and Bottou, L. (eds.),
  \emph{Advances in Neural Information Processing Systems}, volume~21. Curran
  Associates, Inc., 2008.

\bibitem[Johansson et~al.(2016)Johansson, Shalit, and Sontag]{Johansson16}
Johansson, F.~D., Shalit, U., and Sontag, D.
\newblock Learning representations for counterfactual inference.
\newblock In \emph{Proceedings of the 33rd International Conference on
  International Conference on Machine Learning - Volume 48}, ICML'16, pp.\
  3020–3029. JMLR.org, 2016.

\bibitem[LeCun et~al.(2010)LeCun, Cortes, and Burges]{lecun-mnist}
LeCun, Y., Cortes, C., and Burges, C.
\newblock Mnist handwritten digit database.
\newblock \emph{ATT Labs [Online]. Available:
  http://yann.lecun.com/exdb/mnist}, 2, 2010.

\bibitem[Louizos et~al.(2017)Louizos, Shalit, Mooij, Sontag, Zemel, and
  Welling]{Louizos17}
Louizos, C., Shalit, U., Mooij, J., Sontag, D., Zemel, R., and Welling, M.
\newblock Causal effect inference with deep latent-variable models.
\newblock In \emph{Proceedings of the 31st International Conference on Neural
  Information Processing Systems}, NIPS'17, pp.\  6449–6459, Red Hook, NY,
  USA, 2017. Curran Associates Inc.
\newblock ISBN 9781510860964.

\bibitem[Salama(2018)]{keras_vit}
Salama, K.
\newblock Image classification with vision transformer, 2018.
\newblock URL
  \url{https://keras.io/examples/vision/image_classification_with_vision_transformer/}.

\bibitem[Shalit et~al.(2017)Shalit, Johansson, and Sontag]{tarnet}
Shalit, U., Johansson, F.~D., and Sontag, D.
\newblock Estimating individual treatment effect: Generalization bounds and
  algorithms.
\newblock In \emph{Proceedings of the 34th International Conference on Machine
  Learning - Volume 70}, ICML'17, pp.\  3076–3085. JMLR.org, 2017.

\bibitem[Shi et~al.(2019)Shi, Blei, and Veitch]{dragonnet}
Shi, C., Blei, D., and Veitch, V.
\newblock Adapting neural networks for the estimation of treatment effects.
\newblock In Wallach, H., Larochelle, H., Beygelzimer, A., d\textquotesingle
  Alch\'{e}-Buc, F., Fox, E., and Garnett, R. (eds.), \emph{Advances in Neural
  Information Processing Systems}, volume~32. Curran Associates, Inc., 2019.

\bibitem[Vaswani et~al.(2017)Vaswani, Shazeer, Parmar, Uszkoreit, Jones, Gomez,
  Kaiser, and Polosukhin]{transformer}
Vaswani, A., Shazeer, N., Parmar, N., Uszkoreit, J., Jones, L., Gomez, A.~N.,
  Kaiser, L.~u., and Polosukhin, I.
\newblock Attention is all you need.
\newblock In Guyon, I., Luxburg, U.~V., Bengio, S., Wallach, H., Fergus, R.,
  Vishwanathan, S., and Garnett, R. (eds.), \emph{Advances in Neural
  Information Processing Systems}, volume~30. Curran Associates, Inc., 2017.

\end{thebibliography}
\bibliographystyle{bib_style}

\end{document}